\documentclass[runningheads]{llncs}
\usepackage[T1]{fontenc}
\usepackage{graphicx}
\usepackage{amsmath}
\usepackage{algorithm}
\usepackage{algorithmic}
\usepackage{amsfonts}
\usepackage{listings}
\usepackage{xcolor}
\usepackage{hyperref} 
\usepackage{orcidlink}

\definecolor{codegray}{gray}{0.9}
\definecolor{keywordcolor}{RGB}{255, 0, 0}
\definecolor{commentcolor}{RGB}{0, 128, 0}
\definecolor{stringcolor}{RGB}{0, 0, 255}

\begin{document}
\renewcommand{\thefootnote}{\fnsymbol{footnote}}  
\title{Gradient Map-Assisted Head and Neck Tumor Segmentation: A Pre-RT to Mid-RT Approach in MRI-Guided Radiotherapy}
\titlerunning{Gradient Map-Assisted Mid-RT HNC Tumor Segmentation}
%
\author{Jintao Ren \inst{1,2}\orcidID{0000-0002-1558-7196} \and 
Kim Hochreuter \inst{1,2}\orcidID{0009-0000-7347-1688} \and 
Mathis Ersted Rasmussen \inst{1,2}\orcidID{0000-0002-7853-3531} \and 
Jesper Folsted Kallehauge \inst{1,2}\orcidID{0000-0003-3705-5390} \and 
Stine Sofia Korreman\thanks{Corresponding author: stine.korreman@clin.au.dk}\inst{1,2,3}\orcidID{0000-0002-3523-382X} }
\authorrunning{J. Ren et al.}
%
\institute{Aarhus University, Department of Clinical Medicine, Nordre Palle Juul-Jensens Blvd. 11,
8200 Aarhus, Denmark \and
Aarhus University Hospital, Danish Centre for Particle Therapy, Palle Juul-Jensens Blvd. 25, 8200 Aarhus, Denmark \and
Aarhus University, Department of Oncology, Palle Juul-Jensens Blvd. 35,
8200 Aarhus, Denmark \\
\email{stine.korreman@clin.au.dk}
\\ 
}
\maketitle              
\begin{abstract}
    Radiation therapy (RT) is a vital part of treatment for head and neck cancer, where accurate segmentation of gross tumor volume (GTV) is essential for effective treatment planning. This study investigates the use of pre-RT tumor regions and local gradient maps to enhance mid-RT tumor segmentation for head and neck cancer in MRI-guided adaptive radiotherapy. By leveraging pre-RT images and their segmentations as prior knowledge, we address the challenge of tumor localization in mid-RT segmentation. A gradient map of the tumor region from the pre-RT image is computed and applied to mid-RT images to improve tumor boundary delineation. Our approach demonstrated improved segmentation accuracy for both primary GTV (GTVp) and nodal GTV (GTVn), though performance was limited by data constraints. The final DSC$_{agg}$ scores from the challenge's test set evaluation were 0.534 for GTVp, 0.867 for GTVn, and a mean score of 0.70. This method shows potential for enhancing segmentation and treatment planning in adaptive radiotherapy. Team: DCPT-Stine's group.

\end{abstract}
\section{Introduction}
Radiation therapy (RT) is a key treatment modality for head and neck cancer (HNC), but tumor delineation for RT planning remains a major challenge. Traditional RT planning relies heavily on manual delineation of tumor volumes by clinicians, a process that is both time-consuming and prone to high inter-observer variability (IOV), particularly in HNC where complex anatomical structures and critical organs at risk (OARs) lie in close proximity to the tumor \cite{van2019interobserver,nielsen2024interobserver,riegel2006variability}. MRI-guided RT has emerged as a promising approach offering superior soft tissue contrast and the advantage of avoiding additional ionizing radiation during imaging. Further, MRI-guided adaptive RT holds significant potential to improve clinical outcomes by maximizing tumor control while minimizing side effects \cite{benitez2024mri,mohamed2018prospective,mcdonald2024use}. In addition, the use of multimodality images is also recommended \cite{jensen2021imaging}.

With the advancements in imaging techniques and artificial intelligence (AI), recent research has increasingly focused on automating the segmentation of HNC tumors and OARs using deep learning methods \cite{mcdonald2024investigation,rasmussen2023simple}. These AI-driven approaches aim to overcome the limitations of manual segmentation by providing faster and potentially more consistent delineations \cite{hindocha2023artificial,hurkmans2024joint}. However, due to the lack of gold-standard imaging, these applications face a common challenge of uncertainty. Specifically, for HNC, tumors often have ambiguous borders, particularly in the primary tumor volume (GTVp), where the lack of clear distinction between healthy and malignant tissues further complicates tumor delineation for both human annotators and AI models.  Strong IOV is particularly problematic for GTVp, where inconsistent manual annotations can significantly impact the training of deep learning models \cite{henderson2023accurate}.

To further improve tumor segmentation performance for head and neck cancers, previous studies have explored methods that incorporate prior segmentations or prompts (e.g. bounding boxes, scribbles and clicks) to refine subsequent segmentation tasks. For example, Outeiral et al. applied bounding box cropping methods, which led to an increase in DSC for MRI images \cite{outeiral2021oropharyngeal}. Ren et al. found that adding a bounding box as an additional channel improved nnUNet performance, raising GTVp/lymph node (GTVn) DSC from 0.68/0.63 to 0.88/0.89 on multimodal data (CT, PET, MRI) \cite{ren2024segment}. Wang et al. used a RetinaNet model to narrow segmentation fields on CT and PET scans, improving precision \cite{wang2024comparison}. Wei et al. \cite{wei2023towards} demonstrated that incorporating minimal training steps after human interactions raised GTV accuracy from a DSC of 0.65 to 0.82. Interactions such as single or multiple clicks within the GTV have also demonstrated significant improvements in segmentation accuracy \cite{ren2023mo,saukkoriipi2024interactive}.

Building upon these advancements, this study addresses the challenges of MRI-guided adaptive RT for HNC, focusing on the second task of the HNTS-MRG 2024 challenge \cite{wahid2024training}. This task involves segmenting GTVp and GTVn using pre-RT and mid-RT T2-weighted (T2w) MRI images, with mid-RT images taken after RT treatment. The main challenge is accurately identifying all malignant tumor regions and determining the correct contours, complicated by similar soft tissue contrasts and the lack of definitive ground truth \cite{rodriguez2022update}. Tumor shrinkage or disappearance in mid-RT images further complicates boundary delineation.

To overcome these challenges, pre-RT images and their corresponding delineations can serve as valuable prior knowledge for assisting in the segmentation of tumors on mid-RT images. However, the optimal way to leverage this pre-RT information for mid-RT segmentation remains an open question \cite{wang2024prior}.

Despite the temporal link, pre-RT and mid-RT images often differ in intensity, shape, and texture, presenting a challenge for consistent segmentation. These inter- and intra-patient variations may, however, enrich deep learning models by providing diverse training examples for better generalization.

In this study, we present a novel approach that utilizes pre-RT tumor delineations to improve mid-RT segmentation. We first use the deformably registered pre-RT tumor delineations to identify bounding boxes, defining Regions of Interest (ROIs) around the tumors. These ROIs are then employed to compute gradient maps on the mid-RT T2w images, which serve as additional input channels. Additionally, we generate gradient maps from the original pre-RT images and their ground truth (GT) delineations, incorporating them as extra training data. This approach aims to leverage both pre-RT and mid-RT information, thereby enhancing segmentation accuracy during the mid-RT phase.

\section{Material and Methods}
\subsection{Data}
The dataset used in this study was provided by the organizer of the HNTS-MRG 2024 challenge task2 which were 150 HNC patients, predominantly oropharyngeal cancer (OPC). Imaging provided for each patient was T2w fat saturated anatomical sequences of the head and neck region taken at MD Anderson Cancer Center \cite{wahid2024training}. Images include pre-RT (1-3 weeks before the start of RT) and mid-RT (2-4 weeks intra-RT) scans. Multiple physician expert observers (n = 3 to 4) have independently segmented GTVp and GTVn structures for all cases (pre-RT and mid-RT) based on MRI images provided in the challenge. The ground truth was obtained via the Simultaneous Truth And Performance Level Estimation algorithm (STAPLE). Pre-RT images and delineations were deformably registered (DR) to the mid-RT images (DR pre-RT).

\subsection{Incorporating pre-RT Tumor Location with Gradient Maps for mid-RT}
We consider pre-RT images and their delineations as valuable prior knowledge for segmenting tumors on mid-RT images, especially for identifying tumor locations. To leverage this information, we performed connected component analysis on the GTV segmentation masks from DR pre-RT images to identify individual tumor instances (GTVp and GTVn). For each instance, 3D coordinates of the tumor boundaries were extracted, and a bounding box was created around the tumor, with random perturbations of 2-6 voxels in the x, y, and z directions to account for spatial variations. Next, we computed the gradient magnitude (to construct a gradient map) from mid-RT T2w MRI images within these bounding boxes to capture intensity changes around tumor boundaries. The preprocessing involved normalizing the T2w MRI based on its full intensity range and extracting the regions defined by the pre-RT mask. A gradient map was then generated using a Gaussian filter (sigma = 1) to highlight intensity changes. Gradient values were normalized to the range 0-1, with any values greater than 1 clipped to maintain consistency across all patients.

Although pre-RT and mid-RT images are temporally related, their higher-level features, such as tumor morphology and intensity patterns, can differ significantly, likely due to treatment effects or natural variability. To account for this, we treated the pre-RT and mid-RT images of the same patient as independent data points, effectively doubling the dataset size from 150 to 300 samples. Gradient maps were computed for both pre-RT and mid-RT images using the similar process. This approach allowed the model to learn from a broader range of spatial and intensity variations by expanding the training set.

\begin{figure} 
    \includegraphics[width=\textwidth]{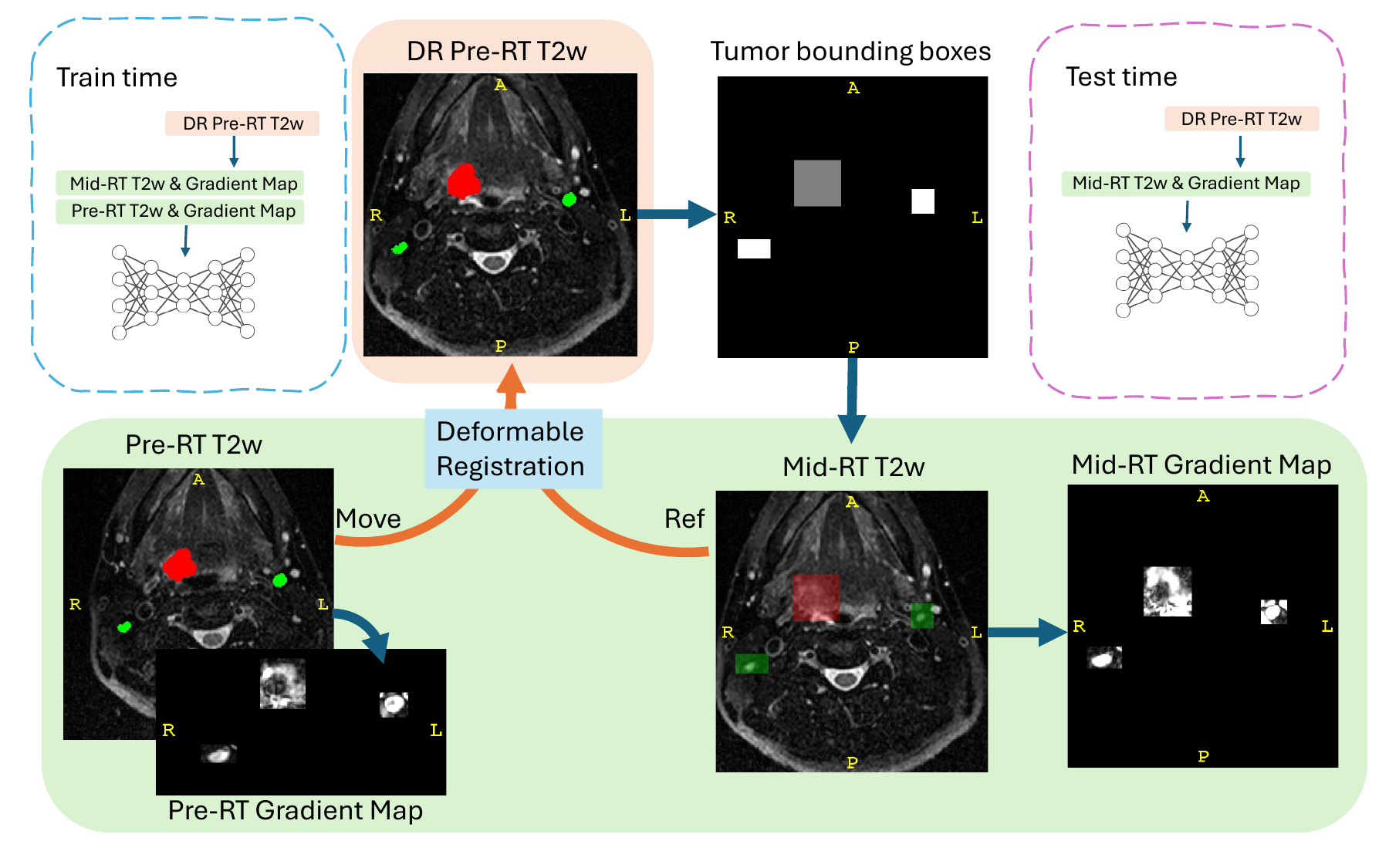}
    \caption{Workflow for incorporating pre-RT tumor location and gradient maps in mid-RT segmentation.}\label{figure1}
    \end{figure}

At test time, the bounding boxes of tumors from DR pre-RT were used to construct the gradient maps of mid-RT T2w MR images. Therefore, both mid-RT T2w and the gradient maps were treated as two-channel inputs for the network. The overall process is outlined in detail in the flowchart, as illustrated in \autoref{figure1}.

\subsection{Deep learning configurations}
We utilized the nnUNet framework \cite{isensee2021nnu} to implement and train the model, employing the nnUNetResEncM planner to design the network architecture\cite{isensee2024nnu}. The model was based on a Residual Encoder U-Net with six stages. Each stage contained a specified number of convolutional blocks: Stage 1 had 1 block, Stage 2 had 3 blocks, Stage 3 had 4 blocks, and Stages 4 through 6 each had 6 blocks. The Conv3D layers employed a kernel size of 1x3x3 for the first layer, followed by 3x3x3 kernels for the subsequent layers. Strides were set to 1x1x1 in the first stage and either 1x2x2 or 2x2x2 in the later stages to reduce dimensionality. InstanceNorm3D was used for normalization, and LeakyReLU served as the activation function.

The training process used a batch size of 8 with a patch size of 48x192x192 voxels. Both the input T2w images and gradient maps were normalized using Z-Score normalization for intensity standardization. Cubic interpolation (order 3) was applied for data resampling, while first-order interpolation was used for segmentation masks. The median image size was 123x512x511 voxels, with voxel spacing set at 1.2x0.5x0.5 mm. Training was performed using the SGD optimizer with a PolyLR scheduler with exponent=0.9, starting with a learning rate of 0.01. The loss function was a combination of cross-entropy and dice loss.

The 150 patients were randomly divided into 5 folds, with each fold containing 240 images/delineations (comprising pre-RT and mid-RT images for 120 patients) for training, while 30 mid-RT images/delineations for validation. The maximum number of epochs was set to 1000 for each model, and the final models from the last epoch were saved for prediction. An ensemble of all models trained across the five folds was used to generate predictions on the test set for the final challenge submission. In accordance with guidelines for reproducibility and verification \cite{hurkmans2024joint}, all source code and trained weights for gradient map generation and deep learning have been made publicly available on GitHub\footnote{\url{https://github.com/Aarhus-RadOnc-AI/GradientSegHNTS}}.

\subsection{Evaluation Metrics}
The HNTS-MRG challenge utilizes the mean aggregated Dice Similarity Coefficient (DSC$_{agg}$) as the primary evaluation metric for ranking \cite{andrearczyk2021overview}. In addition, we also used the conventional Dice Similarity Coefficient (DSC) and the 95th percentile Hausdorff Distance (HD$_{95}$) and mean surface distance (MSD) as supplementary metrics to assess the segmentation performance for both GTVp and GTVn.

\subsection{System Environment}
The experiments were performed on a system featuring dual AMD Ryzen Threadripper 3990X processors with 64 cores (128 threads) and 256GB of RAM. Training was conducted using an NVIDIA RTX A6000 GPU with 48GB of VRAM. The software setup consisted of Python 3.12.4, PyTorch 2.4.0, CUDA 12.6 and nnU-Net 2.5.1, while distance metrics were computed using MedPy 0.5.2.

\section{Results}
\subsection{5-Fold cross-validation results}
We evaluated our approach using 5-fold cross-validation on the training data (n=150). For each fold, average DSC$_{agg}$, HD$_{95}$, and MSD scores were calculated for both GTVp and GTVn, as shown in \autoref{table1} and \autoref{table2}. The results indicated variability in GTVp segmentation accuracy, with DSC$_{agg}$ ranging from 0.469 to 0.697, while GTVn exhibited more consistent performance, with DSC$_{agg}$ ranging from 0.786 to 0.871. The HD$_{95}$ scores ranged from 9.3 to 15.6 mm for GTVp and 4.2 to 7.4 mm for GTVn, while MSD scores varied between 2.5 to 5.4 mm for GTVp and 1.0 to 1.8 mm for GTVn.
\begin{table}
    \caption{GTVp performance on 5-Fold cross-validation}\label{table1}
    \centering
    \begin{tabular}{lcccccc}    
    \hline
    \textbf{Metric} & \textbf{Fold 0} & \textbf{Fold 1} & \textbf{Fold 2} & \textbf{Fold 3} & \textbf{Fold 4} & \textbf{Average} \\
    \hline
    DSC$_{agg}$ & 0.682 & 0.493 & 0.469 & 0.636 & 0.697 & 0.595 \\
    HD$_{95}$ [mm] & 9.9 & 15.6 & 14.6 & 9.3 & 12.3 & 12.3 \\
    MSD [mm] & 3.6 & 5.4 & 4.2 & 2.5 & 3.9 & 3.9 \\
    \hline
    \end{tabular}
    \end{table}
    \begin{table}
    \caption{GTVn performance on 5-Fold cross-validation}\label{table2}
    \centering
     \begin{tabular}{lcccccc}    
    \hline
    \textbf{Metric} & \textbf{Fold 0} & \textbf{Fold 1} & \textbf{Fold 2} & \textbf{Fold 3} & \textbf{Fold 4} & \textbf{Average} \\
    \hline
    DSC$_{agg}$ & 0.871 & 0.786 & 0.868 & 0.859 & 0.827 & 0.842 \\
    HD$_{95}$ [mm] & 4.2 & 5.4 & 6.4 & 7.4 & 5.9 & 5.9 \\
    MSD [mm] & 1.0 & 1.4 & 1.8 & 1.6 & 1.5 & 1.5 \\
    \hline
    \end{tabular}
    \end{table}

\subsection{Comparison between with and without gradient map}
\begin{figure}\centering
    \includegraphics[width=0.6\textwidth]{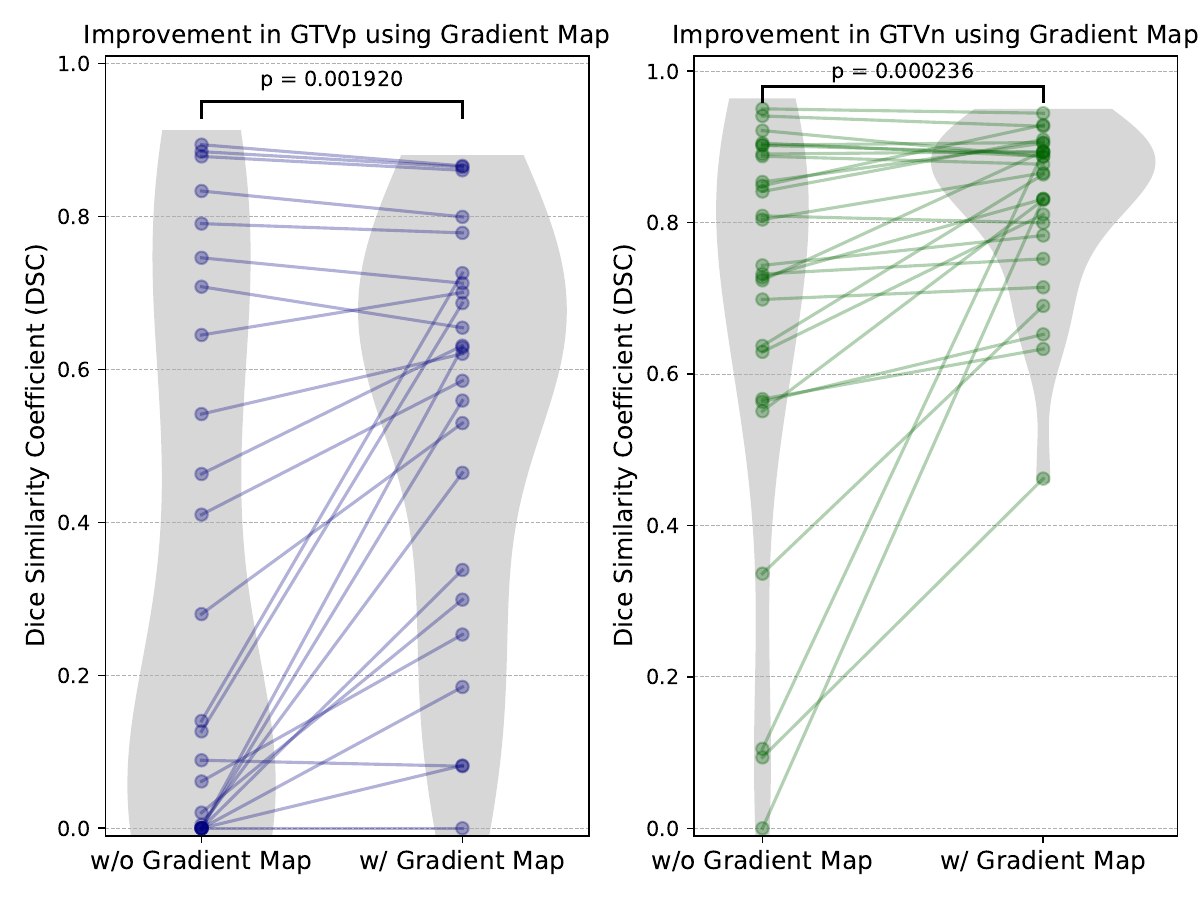}
    \caption{Comparison of segmentation performance with (w/) and without (w/o) gradient maps on fold-0 of the validation set (n=30).}\label{figure2}
    \end{figure}

\begin{figure}\centering
    \includegraphics[width=0.74\textwidth]{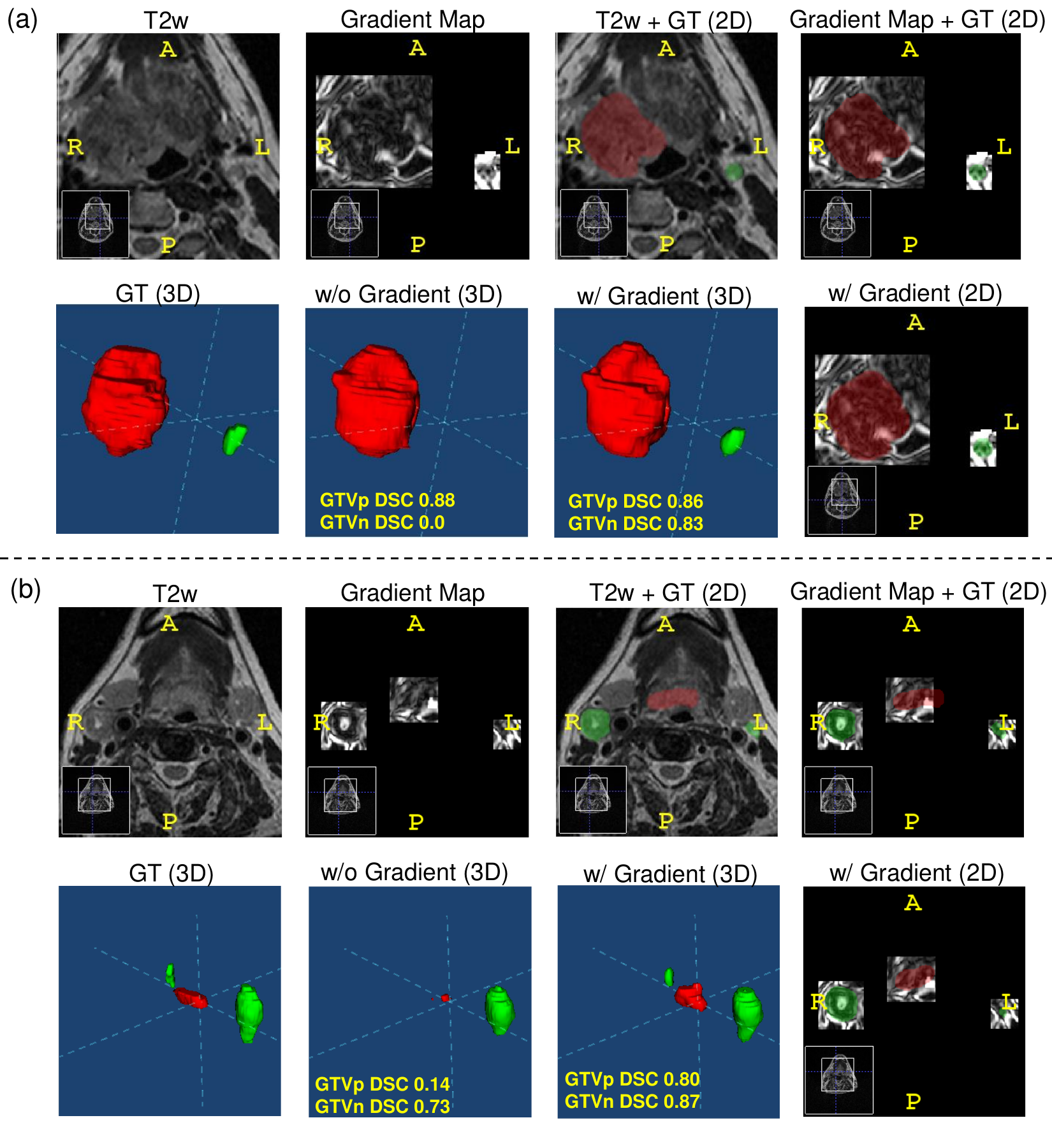}
    \caption{Example cases demonstrating the impact of using gradient maps. \textit{Patient a}: The gradient map enabled accurate segmentation of a previously undetected GTVn, improving the DSC from 0.0 to 0.83. \textit{Patient b}: The use of the gradient map increased the DSC for both GTVp (from 0.14 to 0.80) and GTVn (from 0.73 to 0.87), although part of the GTVp was missed as it extended beyond the gradient map's bounding box.}\label{figure3}
    \end{figure}

We further compared the performance of using T2w images with (w/) and without (w/o) gradient maps on fold-0 of the validation set (n=30). For GTVp, the mean DSC improved from 0.355 to 0.538 (p < 0.005), and for GTVn, it increased from 0.688 to 0.825 (p < 0.001), based on Wilcoxon signed-rank tests. In  \autoref{figure2}, the violin plots illustrate this comparison, showing a clear shift toward higher DSC values for both GTVp and GTVn when the gradient map is applied. The distributions highlight the overall improvement in segmentation accuracy when incorporating gradient maps.

\autoref{figure3} presents two patient cases. In patient a, the use of the gradient map successfully segmented a previously missed GTVn on the T2w-only image, improving the DSC from 0.0 to 0.83. For patient b, both GTVp and GTVn segmentations improved, with DSC scores increasing from 0.14 to 0.80 and 0.73 to 0.87, respectively. However, part of the GTVp was missed as it extended beyond the bounding box range defined by the gradient map.

\subsection{Correlation between tumor volume change and DSC scores}
\begin{figure}[h!]\centering
    \includegraphics[width=\textwidth]{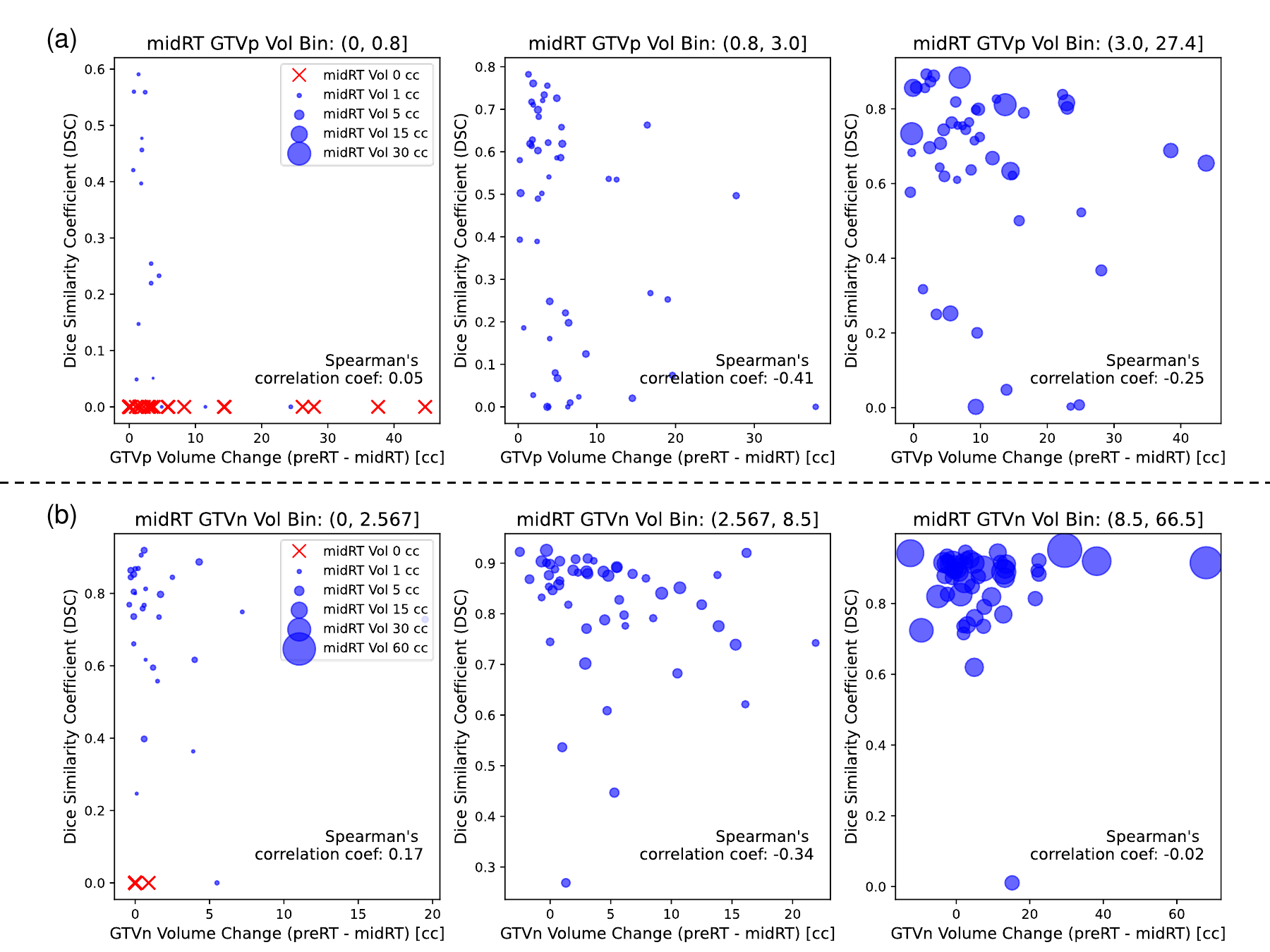}
    \caption{Relationship between tumor volume change and DSC scores for GTVp and GTVn. (a) The first row shows three subplots representing different bins of mid-RT GTVp volumes. Each subplot's x-axis represents the change in volume (pre-RT volume minus mid-RT volume), and the y-axis represents the predicted DSC score. The size of the dots corresponds to the mid-RT GTV volume, while red \textcolor{red}{\textbf{X}} marks indicate instances where the current mid-RT volume is 0. (b) The second row displays similar subplots for GTVn, illustrating the relationship between volume change and segmentation accuracy.}\label{figure4}
    \end{figure}

As shown in \autoref{figure4}, cross-validation results (n=150) revealed no strong correlation between tumor volume change and DSC scores for either GTVp or GTVn. Spearman's correlation coefficients were used to measure these relationships. We observed numerous instances where significant tumor shrinkage resulted in false predictions, particularly for GTVp, with DSC scores of 0.0. Moderate negative correlations were identified in specific volume ranges; for GTVp, a mid-RT volume bin between 0.8 and 3.0 cubic centimeters (cc) showed a Spearman's correlation coefficient of -0.41. Similarly, for GTVn, a volume bin ranging from 2.567 to 8.5 cc had a coefficient of -0.34.

\subsection{Final test score}
After completing training and validation, we submitted our trained model's Docker container to the challenge platform. Predictions were generated using an ensemble of all five models trained across the 5-folds. The final DSC$_{agg}$ scores on the test set, as evaluated by the challenge, were 0.534 for GTVp, 0.867 for GTVn, with a mean score of \textbf{0.70}.

\section{Discussion}
In this study, we integrated pre-RT tumor location and gradient maps to enhance tumor segmentation in mid-RT images for head and neck cancer. Results from the 5-fold cross-validation show that leveraging pre-RT information improved the delineation of both GTVp and GTVn. The average performance across folds, measured by DSC$_{agg}$, HD$_{95}$, and MSD, demonstrated consistent and reliable segmentation. The comparison of  T2w without gradient map versus T2w with gradient map revealed that gradient maps provided more precise boundary localization. These findings highlight the potential of combining spatial and localized gradient intensity information to enhance segmentation accuracy in MRI-guided adaptive radiotherapy.

Deep learning-based AI approaches have garnered significant interest in enhancing RT patient treatment, with notable progress made in HNC tumor auto-segmentation \cite{hindocha2023artificial}. Much of this advancement has been driven by MICCAI public data challenges, such as the HECKTOR challenges \cite{andrearczyk2021overview,Andrearczyk2023}. Numerous studies have participated and contributed diverse approaches to the challenge. Notably, Xie and Peng \cite{xie2021head} achieved a DSC of 0.778 using a 3D SE UNet integrated with the nnUNet pipeline. Similarly, Naser et al. \cite{naser2020tumor} employed a Resnet U-Net, achieving a mean Dice score of 0.69 and further validating the model on MRI, where the mean DSC reached 0.73 ± 0.12 for T2w+ T1w imaging \cite{wahid2022evaluation}.

Despite these advancements, there remains a scarcity of studies or publicly available datasets that leverage the temporal dependency between pre-RT and mid-RT images for head and neck tumor segmentation, especially using MRI data. The HNTS-MRG 2024 Challenge helps address this gap. A primary difficulty in this challenge is accurately identifying all malignant regions in the mid-RT MRI, where using pre-RT ground truth delineations as prior information can help overcome this difficulty. As demonstrated in our study, even a rough propagation of pre-RT information, such as incorporating a bounding box, led to significant improvements in segmentation accuracy. Specifically, the mean DSC for GTVp improved from 0.355 to 0.538 (p < 0.005), and for GTVn, it increased from 0.688 to 0.825 (p < 0.001). These results highlight the potential of utilizing pre-RT data; however, accurately tracking tumor evolution and precisely propagating contours from pre-RT to mid-RT images remains a challenging task, suggesting that more refined methods are needed to fully leverage this temporal information.

In addition to leveraging pre-RT information, our findings indicate that other factors may influence segmentation accuracy. Specifically, in cases where the tumor was absent after treatment (mid-RT GTVp volume = 0), the model failed to predict this absence, resulting in multiple cases of DSC=0.0. This limitation may be exacerbated by the presence of gradient maps in the original tumor region, contributing to false-positive predictions. Furthermore, our results revealed moderate negative correlations in certain volume bins, indicating that as tumors shrink, particularly in the mid-RT moderate volume range GTVp (0.8-3 cc) and GTVn (2.567-8.5 cc), segmentation accuracy tends to decline. This decrease is likely due to the greater complexity in detecting and delineating smaller tumors during the mid-RT phase, where reduced volume leads to greater boundary ambiguity.

Deep learning-based tumor segmentation is inherently data-intensive, and the limited availability of annotated datasets presents a significant challenge. This issue is particularly pronounced in head and neck cancer for mid-RT, where tumor volumes diminish significantly during treatment. Although the dataset consists of 150 training cases, this may be insufficient for advanced techniques like deformation or image propagation networks, which require large datasets to optimize both image alignment and segmentation objectives \cite{kawula2023prior}. Furthermore, variability in patient data distributions complicates training, reducing the model’s ability to generalize across diverse cases.

Our approach aimed to mitigate these challenges by simplifying the problem through a data manipulation approach. Instead of propagating the full image evolution from pre-RT to mid-RT, which would require more complex modeling and a larger dataset, we focused on providing rough tumor localization through the use of bounding boxes derived from pre-RT images. By narrowing the region of interest, we reduced the complexity of the segmentation task. Moreover, we calculated the gradient of the mid-RT image within the bounding box to simplify the training process, allowing more detailed boundary information to be incorporated alongside the location data fed into the network. 

A limitation of our approach is the use of a bounding box with an arbitrary expansion of 2-6 voxels around the deformable registered pre-RT image, which may not always align well with the mid-RT image. This misalignment becomes more critical if there are significant registration errors or substantial anatomical changes in the patient during the course of treatment. The arbitrary choice of expanding the bounding box by 2-6 voxels in all three dimensions carries inherent risks. If the expansion is too small, there is a risk of missing part of the actual tumor in the mid-RT image, as shown on \autoref{figure3}(b). On the other hand, if the expansion is too large, it may reduce the precision of tumor localization, as the box would include more irrelevant tissue, thereby diluting the intended localization benefit.

In routine clinical practice, human involvement is often required to refine such critical tumor definition tasks. Clinicians may define tumor locations interactively, using tools such as scribbles, clicks, or bounding boxes, to provide more accurate localization cues. Studies have demonstrated that integrating such human-in-the-loop methods can significantly enhance detection and segmentation accuracy \cite{saukkoriipi2024interactive,ren2024segment,wei2023towards}. This approach is particularly useful when automated registration struggles with large anatomical changes or complex tumor morphologies, making precise localization difficult without manual input.

Our results demonstrate a significant difference in segmentation accuracy between GTVp and GTVn, with DSC$_{agg}$ scores of 0.534 for GTVp and 0.867 for GTVn. The validation performance for GTVp ranged widely (0.47 to 0.69), whereas GTVn showed more consistency (0.79 to 0.87). This disparity is partly because GTVp often diminishes after treatment, making it barely visible on mid-RT images but still indicated by the gradient map. Additionally, the inherent ambiguity in defining GTVp tumor contours contributes to this difference. The challenge organizers used a STAPLE consensus from multiple annotators (3-4) as the ground truth, but due to low contrast at GTVp boundaries, there may exists considerable IOV, resulting in less agreement and higher uncertainty in GTVp segmentations \cite{segedin2016uncertainties}.

To address this issue, the ground truth delineations should strictly adhere to established guidelines \cite{jensen2020danish}. However, GTVp is often overestimated, leading to a high false positive rate compared to histology, highlighting flaws in delineations and the need for improved imaging techniques \cite{smits2024improved}. From a deep learning developer’s view, an alternative approach could involve developing a probabilistic model based on multiple annotators' input for GTVp, rather than using a consensus method like STAPLE or averaging. Such probabilistic models can significantly improve uncertainty modeling and provide better confident calibration scores, leading to more reliable segmentation maps \cite{ren2024enhancing,kohl2019hierarchical,WAHID2024110542}.

\section{Conclusion}
In conclusion, our novel approach effectively leverages pre-RT information to enhance mid-RT segmentation accuracy by incorporating gradient maps derived from pre-RT tumor delineations. Our results show that utilizing pre-RT delineations improves the model's ability to delineate both GTVp and GTVn for mid-RT, with significant gains in segmentation accuracy. However, the improvements for GTVp are less pronounced, likely due to reduced tumor volume, inherent ambiguity, and variability in tumor contours. We suggest that incorporating a semi-automatic, human-in-the-loop approach could help mitigate false predictions for GTVp, particularly when tumor boundaries are unclear in imaging. Future work could focus on integrating probabilistic models based on multiple annotators to better address this uncertainty. Overall, our approach demonstrates potential for enhancing segmentation accuracy in MRI-guided adaptive radiotherapy.

\bibliographystyle{splncs04}  
\bibliography{reference}  

\begin{thebibliography}{10}
\providecommand{\url}[1]{\texttt{#1}}
\providecommand{\urlprefix}{URL }
\providecommand{\doi}[1]{https://doi.org/#1}

\bibitem{andrearczyk2021overview}
Andrearczyk, V., Oreiller, V., Boughdad, S., Rest, C.C.L., Elhalawani, H.,
  Jreige, M., Prior, J.O., Valli{\`e}res, M., Visvikis, D., Hatt, M., et~al.:
  Overview of the hecktor challenge at miccai 2021: automatic head and neck
  tumor segmentation and outcome prediction in pet/ct images. In: 3D head and
  neck tumor segmentation in PET/CT challenge, pp. 1--37. Springer (2021)

\bibitem{Andrearczyk2023}
Andrearczyk, V., et~al.: Overview of the hecktor challenge at miccai 2022:
  Automatic head and neck tumor segmentation and outcome prediction in pet/ct.
  In: Andrearczyk, V., Oreiller, V., Hatt, M., Depeursinge, A. (eds.) Head and
  Neck Tumor Segmentation and Outcome Prediction. HECKTOR 2022, Lecture Notes
  in Computer Science, vol. 13626. Springer, Cham (2023).
  \doi{10.1007/978-3-031-27420-6_1},
  \url{https://doi.org/10.1007/978-3-031-27420-6_1}

\bibitem{benitez2024mri}
Benitez, C.M., Chuong, M.D., K{\"u}nzel, L.A., Thorwarth, D.: Mri-guided
  adaptive radiation therapy. In: Seminars in Radiation Oncology. vol.~34, pp.
  84--91. Elsevier (2024)

\bibitem{henderson2023accurate}
Henderson, E.G., Osorio, E.M.V., van Herk, M., Brouwer, C.L., Steenbakkers,
  R.J., Green, A.F.: Accurate segmentation of head and neck radiotherapy ct
  scans with 3d cnns: Consistency is key. Physics in Medicine \& Biology
  \textbf{68}(8),  085003 (2023)

\bibitem{hindocha2023artificial}
Hindocha, S., Zucker, K., Jena, R., Banfill, K., Mackay, K., Price, G., Pudney,
  D., Wang, J., Taylor, A.: Artificial intelligence for radiotherapy
  auto-contouring: Current use, perceptions of and barriers to implementation.
  Clinical Oncology  \textbf{35}(4),  219--226 (2023)

\bibitem{hurkmans2024joint}
Hurkmans, C., Bibault, J.E., Brock, K.K., van Elmpt, W., Feng, M., Fuller,
  C.D., Jereczek-Fossa, B.A., Korreman, S., Landry, G., Madesta, F., et~al.: A
  joint estro and aapm guideline for development, clinical validation and
  reporting of artificial intelligence models in radiation therapy.
  Radiotherapy and Oncology  \textbf{197},  110345 (2024)

\bibitem{isensee2021nnu}
Isensee, F., Jaeger, P.F., Kohl, S.A., Petersen, J., Maier-Hein, K.H.: nnu-net:
  a self-configuring method for deep learning-based biomedical image
  segmentation. Nature methods  \textbf{18}(2),  203--211 (2021)

\bibitem{isensee2024nnu}
Isensee, F., Wald, T., Ulrich, C., Baumgartner, M., Roy, S., Maier-Hein, K.,
  Jaeger, P.F.: nnu-net revisited: A call for rigorous validation in 3d medical
  image segmentation. arXiv preprint arXiv:2404.09556  (2024)

\bibitem{jensen2021imaging}
Jensen, K., Al-Farra, G., Dejanovic, D., Eriksen, J.G., Loft, A., Hansen, C.R.,
  Pameijer, F.A., Zukauskaite, R., Grau, C.: Imaging for target delineation in
  head and neck cancer radiotherapy. In: Seminars in nuclear medicine. vol.~51,
  pp. 59--67. Elsevier (2021)

\bibitem{jensen2020danish}
Jensen, K., Friborg, J., Hansen, C.R., Sams{\o}e, E., Johansen, J., Andersen,
  M., Smulders, B., Andersen, E., Nielsen, M.S., Eriksen, J.G., et~al.: The
  danish head and neck cancer group (dahanca) 2020 radiotherapy guidelines.
  Radiotherapy and Oncology  \textbf{151},  149--151 (2020)

\bibitem{kawula2023prior}
Kawula, M., Vagni, M., Cusumano, D., Boldrini, L., Placidi, L., Corradini, S.,
  Belka, C., Landry, G., Kurz, C.: Prior knowledge based deep learning
  auto-segmentation in magnetic resonance imaging-guided radiotherapy of
  prostate cancer. Physics and Imaging in Radiation Oncology  \textbf{28},
  100498 (2023)

\bibitem{kohl2019hierarchical}
Kohl, S.A., Romera-Paredes, B., Maier-Hein, K.H., Rezende, D.J., Eslami, S.,
  Kohli, P., Zisserman, A., Ronneberger, O.: A hierarchical probabilistic u-net
  for modeling multi-scale ambiguities. arXiv preprint arXiv:1905.13077  (2019)

\bibitem{mcdonald2024investigation}
McDonald, B.A., Cardenas, C.E., O'Connell, N., Ahmed, S., Naser, M.A., Wahid,
  K.A., Xu, J., Thill, D., Zuhour, R.J., Mesko, S., et~al.: Investigation of
  autosegmentation techniques on t2-weighted mri for off-line dose
  reconstruction in mr-linac workflow for head and neck cancers. Medical
  physics  \textbf{51}(1),  278--291 (2024)

\bibitem{mcdonald2024use}
McDonald, B.A., Dal~Bello, R., Fuller, C.D., Balermpas, P.: The use of
  mr-guided radiation therapy for head and neck cancer and recommended
  reporting guidance. In: Seminars in radiation oncology. vol.~34, pp. 69--83.
  Elsevier (2024)

\bibitem{mohamed2018prospective}
Mohamed, A.S., Bahig, H., Aristophanous, M., Blanchard, P., Kamal, M., Ding,
  Y., Cardenas, C.E., Brock, K.K., Lai, S.Y., Hutcheson, K.A., et~al.:
  Prospective in silico study of the feasibility and dosimetric advantages of
  mri-guided dose adaptation for human papillomavirus positive oropharyngeal
  cancer patients compared with standard imrt. Clinical and translational
  radiation oncology  \textbf{11},  11--18 (2018)

\bibitem{naser2020tumor}
Naser, M.A., van Dijk, L.V., He, R., Wahid, K.A., Fuller, C.D.: Tumor
  segmentation in patients with head and neck cancers using deep learning
  based-on multi-modality pet/ct images. In: 3D Head and Neck Tumor
  Segmentation in PET/CT Challenge, pp. 85--98. Springer (2020)

\bibitem{nielsen2024interobserver}
Nielsen, C.P., Lorenzen, E.L., Jensen, K., Eriksen, J.G., Johansen, J.,
  Gyldenkerne, N., Zukauskaite, R., Kjellgren, M., Maare, C., L{\o}nkvist,
  C.K., et~al.: Interobserver variation in organs at risk contouring in head
  and neck cancer according to the dahanca guidelines. Radiotherapy and
  Oncology  \textbf{197},  110337 (2024)

\bibitem{outeiral2021oropharyngeal}
Outeiral, R.R., Bos, P., Al-Mamgani, A., Jasperse, B., Sim{\~o}es, R., van~der
  Heide, U.A.: Oropharyngeal primary tumor segmentation for radiotherapy
  planning on magnetic resonance imaging using deep learning. Physics and
  imaging in radiation oncology  \textbf{19},  39--44 (2021)

\bibitem{rasmussen2023simple}
Rasmussen, M.E., Nijkamp, J.A., Eriksen, J.G., Korreman, S.S.: A simple
  single-cycle interactive strategy to improve deep learning-based segmentation
  of organs-at-risk in head-and-neck cancer. Physics and Imaging in Radiation
  Oncology  \textbf{26},  100426 (2023)

\bibitem{ren2023mo}
Ren, J., Nijkamp, J., Rasmussen, M., Eriksen, J., Korreman, S.: Mo-0799
  single-click user input reduces false detection in deep learning head and
  neck tumor segmentation. Radiotherapy and Oncology  \textbf{182},  S669--S671
  (2023)

\bibitem{ren2024segment}
Ren, J., Rasmussen, M., Nijkamp, J., Eriksen, J.G., Korreman, S.: Segment
  anything model for head and neck tumor segmentation with ct, pet and mri
  multi-modality images. arXiv preprint arXiv:2402.17454  (2024)

\bibitem{ren2024enhancing}
Ren, J., Teuwen, J., Nijkamp, J., Rasmussen, M., Gouw, Z., Eriksen, J.G.,
  Sonke, J.J., Korreman, S.: Enhancing the reliability of deep learning-based
  head and neck tumour segmentation using uncertainty estimation with
  multi-modal images. Physics in Medicine \& Biology  \textbf{69}(16),  165018
  (2024)

\bibitem{riegel2006variability}
Riegel, A.C., Berson, A.M., Destian, S., Ng, T., Tena, L.B., Mitnick, R.J.,
  Wong, P.S.: Variability of gross tumor volume delineation in head-and-neck
  cancer using ct and pet/ct fusion. International Journal of Radiation
  Oncology* Biology* Physics  \textbf{65}(3),  726--732 (2006)

\bibitem{rodriguez2022update}
Rodriguez, J.D., Selleck, A.M., Razek, A.A.K.A., Huang, B.Y.: Update on mr
  imaging of soft tissue tumors of head and neck. Magnetic Resonance Imaging
  Clinics  \textbf{30}(1),  151--198 (2022)

\bibitem{saukkoriipi2024interactive}
Saukkoriipi, M., Sahlsten, J., Jaskari, J., Orasmaa, L., Kangas, J., Rasouli,
  N., Raisamo, R., Hirvonen, J., Mehtonen, H., J{\"a}rnstedt, J., et~al.:
  Interactive 3d segmentation for primary gross tumor volume in oropharyngeal
  cancer. arXiv preprint arXiv:2409.06605  (2024)

\bibitem{segedin2016uncertainties}
Segedin, B., Petric, P.: Uncertainties in target volume delineation in
  radiotherapy--are they relevant and what can we do about them? Radiology and
  oncology  \textbf{50}(3),  254--262 (2016)

\bibitem{smits2024improved}
Smits, H.J., Raaijmakers, C.P., de~Ridder, M., Gouw, Z.A., Doornaert, P.A.,
  Pameijer, F.A., Lodeweges, J.E., Ruiter, L.N., Kuijer, K.M., Schakel, T.,
  et~al.: Improved delineation with diffusion weighted imaging for laryngeal
  and hypopharyngeal tumors validated with pathology. Radiotherapy and Oncology
   \textbf{194},  110182 (2024)

\bibitem{van2019interobserver}
van~der Veen, J., Gulyban, A., Nuyts, S.: Interobserver variability in
  delineation of target volumes in head and neck cancer. Radiotherapy and
  Oncology  \textbf{137},  9--15 (2019)

\bibitem{wahid2024training}
Wahid, K., Dede, C., Naser, M., Fuller, C.: Training dataset for hntsmrg 2024
  challenge. \url{https://doi.org/10.5281/zenodo.11199559} (2024), [Data set]

\bibitem{wahid2022evaluation}
Wahid, K.A., Ahmed, S., He, R., van Dijk, L.V., Teuwen, J., McDonald, B.A.,
  Salama, V., Mohamed, A.S., Salzillo, T., Dede, C., et~al.: Evaluation of deep
  learning-based multiparametric mri oropharyngeal primary tumor
  auto-segmentation and investigation of input channel effects: Results from a
  prospective imaging registry. Clinical and translational radiation oncology
  \textbf{32},  6--14 (2022)

\bibitem{WAHID2024110542}
Wahid, K.A., Kaffey, Z.Y., Farris, D.P., Humbert-Vidan, L., Moreno, A.C.,
  Rasmussen, M., Ren, J., Naser, M.A., Netherton, T.J., Korreman, S.,
  Balakrishnan, G., Fuller, C.D., Fuentes, D., Dohopolski, M.J.: Artificial
  intelligence uncertainty quantification in radiotherapy applications - a
  scoping review. Radiotherapy and Oncology  \textbf{201},  110542 (2024).
  \doi{https://doi.org/10.1016/j.radonc.2024.110542},
  \url{https://www.sciencedirect.com/science/article/pii/S0167814024035205}

\bibitem{wang2024prior}
Wang, X., Chang, Y., Pei, X., Xu, X.G.: A prior-information-based automatic
  segmentation method for the clinical target volume in adaptive radiotherapy
  of cervical cancer. Journal of Applied Clinical Medical Physics
  \textbf{25}(5),  e14350 (2024)

\bibitem{wang2024comparison}
Wang, Y., Lombardo, E., Huang, L., Avanzo, M., Fanetti, G., Franchin, G.,
  Zschaeck, S., Weing{\"a}rtner, J., Belka, C., Riboldi, M., et~al.: Comparison
  of deep learning networks for fully automated head and neck tumor delineation
  on multi-centric pet/ct images. Radiation Oncology  \textbf{19}(1), ~3 (2024)

\bibitem{wei2023towards}
Wei, Z., Ren, J., Korreman, S.S., Nijkamp, J.: Towards interactive
  deep-learning for tumour segmentation in head and neck cancer radiotherapy.
  Physics and Imaging in Radiation Oncology  \textbf{25},  100408 (2023)

\bibitem{xie2021head}
Xie, J., Peng, Y.: The head and neck tumor segmentation based on 3d u-net. In:
  3D Head and Neck Tumor Segmentation in PET/CT Challenge, pp. 92--98. Springer
  (2021)

\end{thebibliography}
\end{document}